\documentclass[11pt]{article}

\usepackage[final]{acl}

\usepackage{times}
\usepackage{latexsym}

\usepackage[T1]{fontenc}

\usepackage[utf8]{inputenc}

\usepackage{microtype}

\usepackage{inconsolata}

\usepackage{graphicx}

\usepackage{booktabs}
\title{DermaVQA-DAS: Dermatology Assessment Schema (DAS) \& Datasets for Closed-Ended Question Answering And Segmentation}

\author{ Wen{-}wai Yim\thanks{Corresponding Author: Wen{-}wai Yim, email: yimwenwai@microsoft.com}\\
  Microsoft AI, Mountain View, USA\\
  \texttt{yimwenwai@microsoft.com} \\  \And
  Yujuan Fu \\
  University of Washington, Seattle, USA \\
  \texttt{velvinfu@uw.edu} \\ \AND
Asma Ben Abacha\\
  Microsoft AI, Mountain View, USA\\
  \texttt{abenabacha@microsoft.com} \\ \And
  Meliha Yetisgen \\
  University of Washington, Seattle, USA \\
  \texttt{melihay@uw.edu} \\  \AND
  Noel Codella\\
  Microsoft AI, Mountain View, USA\\ 
  \texttt{ncodella@microsoft.com} \\ \AND
  Roberto Andres Novoa \\
  Stanford University, \\
  Stanford, USA \\
  \texttt{rnovoa@stanford.edu}\\\And
  Josep Malvehy \\
  Hospital Clinic of Barcelona,\\
  Barcelona, Spain \\
  \texttt{jmalvehy@clinic.cat}
}

\begin{document}
\maketitle
\begin{abstract}
Recent advances in dermatological image analysis have been driven by large-scale annotated datasets; however, most existing benchmarks focus on dermatoscopic images and lack patient-authored queries and broader clinical context. To address this gap, we introduce the DermaVQA-DAS dataset that supports two complementary tasks: closed-ended question answering (QA) and dermatological lesion segmentation. Central to this work, we introduce the Dermatology Assessment Schema (DAS), a novel expert-developed framework that systematically captures clinically meaningful dermatological features in a structured and standardized format across 72 high-level and 137 fine-grained assessment questions, with multiple-choice options in English and Chinese. For segmentation, we evaluate multiple prompting strategies, identifying that augmented prompts incorporating both the patient query title and content yields the highest performance under majority-vote-based microscore evaluation, achieving a Jaccard index of 0.395 and a Dice score of 0.566 with BiomedParse. For closed-ended QA, overall performance is consistently strong across models, with average accuracies ranging from 0.729 to 0.798; o3 achieves the highest overall accuracy (0.798), closely followed by GPT-4.1 (0.796), while Gemini-1.5-Pro demonstrates competitive performance within the Gemini family (0.783).
We publicly release DermaVQA-DAS, the DAS schema, and standardized evaluation protocols to support and accelerate future research in patient-centered dermatological vision-language modeling (URL) which was the dataset used for international challenge(URL)
\end{abstract}

\begin{figure*}[h]
\begin{center}
\includegraphics[scale=.3]{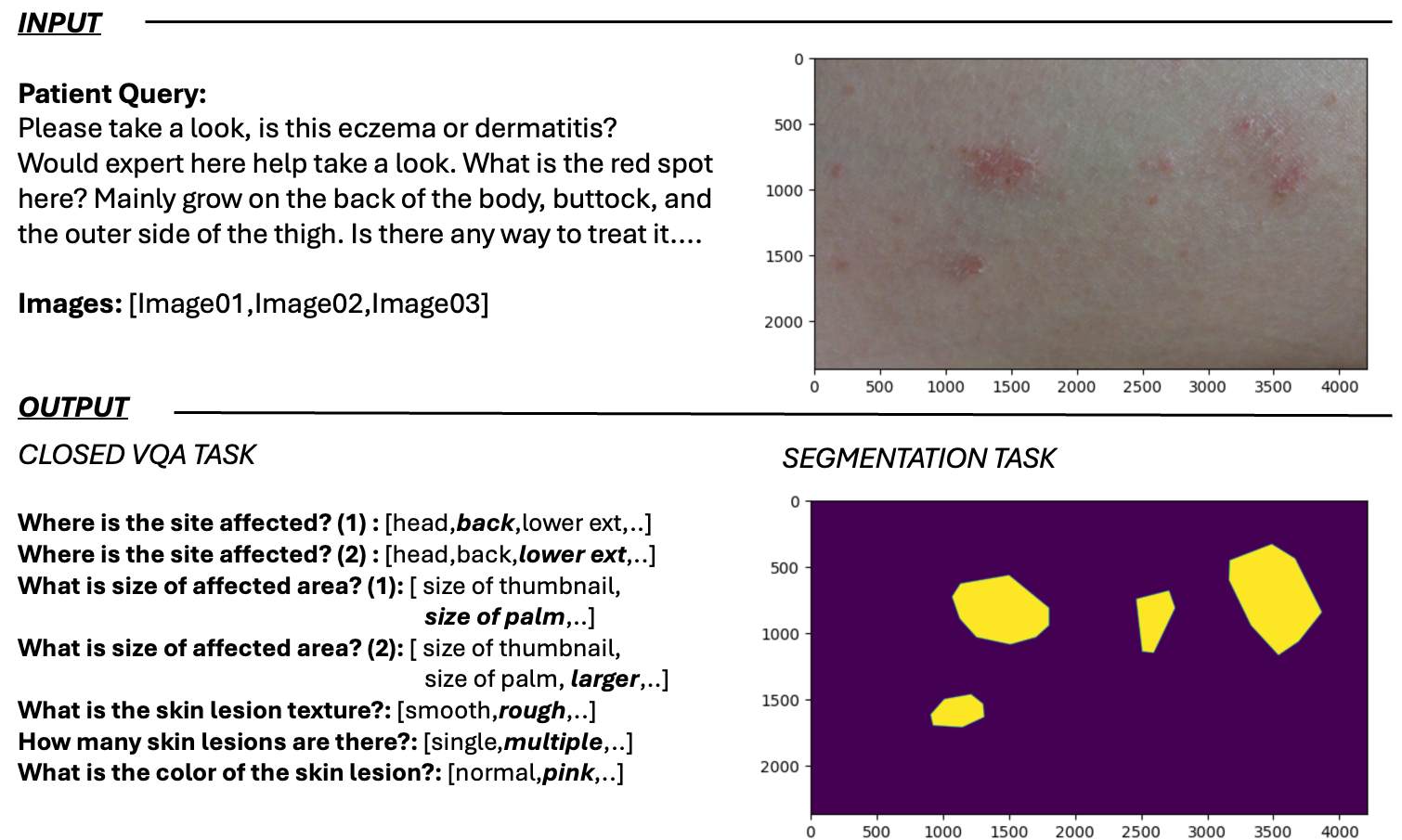}
\caption{An example from the DermaVQA-DAS dataset with the original consumer health query and the accompanying images, as well as the created closed-ended question-answer pairs and segmentation annotations. } 
\label{fig:queryex}
\end{center}
\end{figure*} 

\label{sec:intro}

The launch of the ISIC challenges in 2016 catalyzed progress in dermatological image classification by providing large annotated datasets \cite{ISIC-2016}. However, these datasets focus primarily on dermatoscopic images and lack patient queries and clinical histories, limiting their applicability to patient-centered clinical settings.

To address this gap, prior work introduced DermaVQA~\cite{dermaVQA-2024}, the first dataset combining patient-generated dermatological images with patient queries and free-text expert responses. DermaVQA has been used in the MEDIQA-MAGIC 2024~\cite{mediqa-magic-2024} and MEDIQA-M3G~\cite{mediqa-m3g-2024} competitions. Subsequently, WoundcareVQA~\cite{woundcareVQA-2025} was released, extending visual question answering (VQA) to the domain of wound care. Despite these advances, existing dermatological VQA datasets primarily support open-ended answer generation and lack structured, clinically grounded assessment tasks that enable standardized evaluation and closer alignment with real-world diagnostic workflows. 

This work extends the DermaVQA collection by introducing DermaVQA-DAS, a dataset designed to support two new tasks: closed-ended question answering (QA) and dermatological lesion segmentation. Closed-ended QA enables objective evaluation, reduces ambiguity inherent in free-text generation, and facilitates reliable benchmarking across models. A key contribution is the Dermatology Assessment Schema (DAS), a novel, patient-facing expert-developed framework for structured dermatological assessment based on clinically relevant features. By formalizing dermatological evaluation into structured and hierarchically organized components, DAS supports systematic multimodal reasoning and improves interpretability in clinically relevant settings. DAS comprises 72 high-level and 137 fine-grained questions, each with multiple-choice options available in both English and Chinese. Experiments focus on the nine most frequently populated query questions and their corresponding 27 fine-grained questions. We publicly release the dataset along with the DAS schema to support future research.  

Several multimodal models are benchmarked on the closed-ended QA and segmentation tasks, evaluated under settings both with and without access to clinical history.  This evaluation setup introduces a challenging multimodal reasoning problem that requires integrating visual evidence with structured clinical context. The dataset here was used in the XXXX international shared task challenge 2026. Figure~\ref{fig:queryex} illustrates an example from DermaVQA-DAS, showing the original patient query and images alongside the derived closed-ended QA pairs and corresponding segmentation annotations. \underline{The main contributions of this work are as follows:}

\begin{itemize}
    \item The introduction of DermaVQA-DAS, extending prior dermatological VQA datasets with structured closed-ended QA and segmentation tasks, in Chinese and English.
    \item The Dermatology Assessment Schema (DAS), a hierarchically organized framework comprising 72 high-level and 137 fine-grained clinically grounded questions. Question/answers are designed to be understandable by consumers.
    \item Benchmark performances of segmentation (with and without textual clinical history) and multimodal model performances closed question answering.
    \item Public release of the dataset and schema to facilitate reproducible research in multimodal dermatology.
\end{itemize}


\begin{table*}[h]
\centering
\caption{Question Relevance By Encounter Percentage (Top 20)}
\label{tab:questionrelevance}
\small
\begin{tabular}{c c p{7cm} c}
\hline
\textbf{Category} & \textbf{Type} & \textbf{Question} & \textbf{Percent} \\
\hline
General & Site & How much of the body is effected? & 0.99 \\
General & Site & 1 Where is the affected area? & 0.98 \\
Skin Specific & Affected area description & 1 What label best describes the affected area? & 0.98\\
Skin Specific & Lesion Color & Compared to the normal surrounding skin, what is the color of the skin lesion? & 0.97 \\
Skin Specific & Lesion Count & How many skin lesions are there? & 0.94 \\
General & Size & 1 How large are the affected areas? Please specify which affected area for each selection. & 0.94 \\
General & Onset & When did the patient first notice the issue? & 0.60 \\
Skin Specific & Itch & Is there any associated itching with the skin problem? & 0.54 \\
Skin Specific & Lesion Texture & What is the skin lesion texture? & 0.54 \\
General & Sex & What is the sex of the patient? & 0.51 \\
General & Age & How old is the patient? & 0.42 \\
General & Site & 2 Where is the affected area? & 0.26 \\
Skin Specific & Discharge & Is there associated discharge with the skin condition? & 0.23 \\
Skin Specific & Pain & Is there any associated pain with the skin problem? & 0.21 \\
Skin Specific & Itch & If yes, on a scale of 0-4, with 0 being no itching and 4 being very severe itching, how severe is the itching? (please interpret other descriptive terms for severity, such as "slightly" and "very" to this scale) & 0.21 \\
Skin Specific & Affected area description & 2 What label best describes the affected area?	0.19 \\
Skin Specific & Bleeding & Is there associated bleeding with the skin condition? & 0.19 \\
General & Size & 2 How large are the affected areas? Please specify which affected area for each selection. & 0.18 \\
Time Course & Skin evolution & Has the symptoms spread over time? & 0.17 \\
Time Course & If Rash & Has the rash changed over time? & 0.17 \\
\hline
\end{tabular}
\end{table*}


\section{Related Work}

Large-scale benchmarks such as the ISIC challenges \cite{ISIC-2016,isic-2020,isic-2024} have driven substantial progress in dermatological image classification and lesion segmentation using dermatoscopic images.  However, these datasets focus on curated images with structured diagnostic labels and do not incorporate patient queries or clinical context, thereby limiting evaluation of multimodal reasoning in patient-centered scenarios. Specifically, highly performant models such as MedSAM, trained and evaluated on a variety of medical images including X-Ray, MRI, CT, etc does not allow textual prompting, nor focuses on patient generated mobile phone color images\cite{MedSAM}. BiomedParse features medical image prompt-based joint segmentation and textual description generation, however while dermatoscopy images are trained and evaluated; consumer-generated dermatological images were not included\cite{zhao2025foundation,zhao2025boltzmann}. Here we provide 3 segmentation annotations per image for such consumer-generated images with uncontrolled angle, lighting, and resolutions.

A majority of dermatological consumer-health image datasets work revolve around classification\cite{groh2021evaluatingdeepneuralnetworks,Daneshjou2021DisparitiesID,skincon,sun2016benchmark,JALABOI2023102647,9671333}. Chiefly such datasets are typically labeled by clinicians along highly technical language for diagnoses (e.g. psoriasis), malignancies, and dermatological features (e.g. lichenification). In this work, unlike the former, we pose a schema aiming to be patient-facing and understandable to lay persons. Closest to this approach was the SCIN dataset created by Google\cite{ward2024creating}. To acquire their dataset, they recruited voluntary dermatological image submissions through ads, with additional survey of user supplied information (e.g. onset duration). Unlike this work, our dataset includes natural patient textual queries.

Medical visual question answering (VQA) has emerged as a framework for integrating image understanding and language reasoning. Early benchmarks such as VQA-RAD \cite{lau2018vqarad}, VQA-Med \cite{vqa-med-2019,vqa-med-2020,vqa-med-2021}, and PathVQA \cite{he2020pathvqa} established multimodal QA settings in radiology and pathology. In dermatology, DermaVQA~\cite{dermaVQA-2024}, used in the MEDIQA-MAGIC 2024~\cite{mediqa-magic-2024} and MEDIQA-M3G~\cite{mediqa-m3g-2024} challenges, introduced patient-generated images paired with free-text questions and expert responses, while WoundcareVQA~\cite{woundcareVQA-2025} extended this paradigm to wound assessment. Nevertheless, existing dermatological VQA datasets primarily emphasize open-ended answer generation, limiting standardized evaluation and structured clinical interpretability.

Although lesion segmentation has been extensively studied as a standalone task, it has rarely been integrated with structured question answering in a unified benchmark. In contrast, our work introduces a hierarchically organized Dermatology Assessment Schema (DAS) and supports both closed-ended QA and lesion segmentation within a single dataset, enabling systematic evaluation of clinically grounded multimodal reasoning with and without clinical history.

\begin{table*}[h]
\centering
\caption{Extract of the Dermatology Assessment Schema (DAS). 
Repeated slots enable representation of multiple affected sites or lesions. 
Example options are illustrative rather than exhaustive.}
\label{tab:das_summary}
\small
\begin{tabular}{llccc}
\toprule
CQID & Question Type & Slots & \# Options & Example Options \\
\midrule
CQID010 & Site Extent & 1 & 4 & single spot; widespread \\
CQID011 & Site Location & 6 & 8 & head; upper extremities \\
CQID012 & Size & 6 & 4 & thumb nail; palm \\
CQID015 & Onset & 1 & 7 & within days; within months \\
\midrule
CQID020 & Skin Description & 6 & 10 & raised; flat; crust \\
CQID025 & Itch & 1 & 3 & yes; no \\
CQID034 & Lesion Color & 1 & 12 & red; brown; hyperpigmentation \\
CQID035 & Lesion Count & 1 & 3 & single; multiple \\
CQID036 & Texture & 1 & 3 & smooth; rough \\
\bottomrule
\end{tabular}
\end{table*}

\begin{table}[h]
\centering
\caption{Segmentation dataset statistics across splits.}
\label{tab:segmentation_splits}
\small
\begin{tabular}{lccc}
\toprule
\textbf{Split} & \textbf{Encounters} & \textbf{Images} & \textbf{Masks} \\
\midrule
Train & 842 & 2,474 & 7,448 \\
Valid & 56 & 157 & 472 \\
Test  & 100 & 314 & 944 \\
\midrule
\textbf{Total} & \textbf{998} & \textbf{2,945} & \textbf{8,864} \\
\bottomrule
\end{tabular}
\end{table}

\section{Dataset Creation}

We build upon the publicly available DermaVQA dataset \cite{dermaVQA-2024} to introduce additional structured and vision-centric dermatological reasoning tasks. The resulting dataset, DermaVQA-DAS\footnote{\url{URL\_hidden\_for\_review}}, introduces two new tasks: closed-ended question answering (QA) and dermatological lesion segmentation. 

To enable these tasks, we develop the Dermatology Assessment Schema (DAS), the first open, patient-facing, expert-developed framework for systematically identifying clinically relevant dermatological features. The motivation for this is was to enable a framework in which clinically relevant information, expressed in non-clinical terms, can be captured. DAS was created by two board-certified dermatologists and consists of 72 high-level questions, each associated with multiple-choice answer options. To ensure comprehensibility by non clinicians, two college graduates reviewed question and answers.

To account for multiple sites of involvement, certain questions include duplicate slots (e.g., ``Anatomic Location of Problem (Location 1)'', ``Anatomic Location of Problem (Location 2)'', ``Size at Location 1)''), resulting in a total of 137 fine-grained questions. Consistent with clinical documentation practice, duplicate fields are left empty when not applicable. To understand which questions are relevant for the majority of encounters, we tallied the percentage of encounters populated per question. Table \ref{tab:questionrelevance} provides the top 20 questions, ranked by the highest relevance. To promote accessibility, all questions and answer choices are provided in both English and Chinese. Results are reported for the subset of question categories in which the test and validation splits were populated for at least 51\% of encounters, amounting to nine overall categories comprising 27 closed-ended questions. An extract of the abbreviated Dermatology Assessment Schema (DAS) benchmarked in this paper is shown in Table~\ref{tab:das_summary}.

To measure inter-rater variability we assess 21 encounters annotated by the three annotators across both accuracy and f1 detection and label concordance. Measuring both accuracy and f1 was necessary as we noted that many question are left blank when multiple sites are not applicable (e.g. ``Size at Location 2''). Thus, its possible for accuracy can be artificially high as most blank answers agreed. Detection f1 and accuracy (whether the same questions were answered or left blank due to multiple findings/complaints per encounter) was found to be 0.73 and 0.95, whereas label f1 and accuracy was found to be 0.51 and 0.93. To confirm annotation stability, we found that 88\% of fine-grained labels had clear 2 labeler majority agreement. The full corpus was annotated by three medical annotators with the majority vote as the gold standard. 

The closed-ended QA dataset contains 300 training, 56 validation, and 100 test instances. 
Segmentation dataset statistics are summarized in Table~\ref{tab:segmentation_splits}. 
Across splits, the segmentation dataset contains 998 encounters paired with 2,945 images and 8,864 expert-annotated masks, marking clinically relevant regions of interest. 
Each image contains three expert-provided masks, annotated by four different medical annotators.

\begin{table*}[h!]
\centering
\caption{Segmentation performance under different prompting strategies. Prompt I (default: \textit{"Highlight the abnormal skin lesion that appears different from surrounding healthy skin."}), Prompt II (prompt augmented with patient query title), and Prompt III (prompt augmented with query title and content). Best results within each aggregation scheme are shown in bold.}
\label{tab:segmentation-res}
\begin{tabular}{lcccccc}
\toprule
 & \multicolumn{2}{c}{Mean of Max} 
 & \multicolumn{2}{c}{Mean of Mean} 
 & \multicolumn{2}{c}{Majority Vote} \\
\cmidrule(lr){2-3}
\cmidrule(lr){4-5}
\cmidrule(lr){6-7}
\textbf{Model} 
 & Jaccard & Dice 
 & Jaccard & Dice 
 & Jaccard & Dice \\
\midrule
MedSAM (no prompt) 
 & 0.4097 & 0.5250 
 & 0.3299 & 0.4386 
 & 0.2597 & 0.4123 \\

\midrule
BiomedParse (prompt I) 
 & \bf 0.5088 & \bf 0.6133 
 & \bf 0.4253 & \bf 0.5315 
 & 0.3863 & 0.5573 \\

BiomedParse (prompt II) 
 & 0.5034 & 0.6068 
 & 0.4215 & 0.5268 
 & 0.3893 & 0.5604 \\

BiomedParse (prompt III) 
 & 0.5081 & 0.6120 
 & 0.4243 & 0.5298 
 & \bf 0.3948 & \bf 0.5661 \\
\bottomrule
\end{tabular}
\end{table*}


\section{Methods}

\subsection{Image Segmentation}

For the dermatological lesion segmentation task, we build two methods based on the BiomedParse and MedSAM\footnote{We use a pre-trained MedSAM model hosted on Hugging Face for deployment convenience: \url{https://huggingface.co/flaviagiammarino/MedSAM-vit-base}} models. 
We explore four prompting conditions: 
\begin{enumerate}
    \item no prompt (MedSAM only),
    \item a default prompt ("Highlight the abnormal skin lesion that appears different from surrounding healthy skin."),
    \item a prompt augmented with the patient query title, and
    \item a prompt including both the title and the full query context.
\end{enumerate}

To determine an optimal segmentation threshold, we perform a grid search over 50 evenly spaced threshold values between 0 and 1 using the validation set, selecting the threshold that maximizes the Dice score, and apply the selected threshold to the test set. 

\subsection{Visual Question Answering}

For visual question answering (VQA) in dermatology, each question-answer instance consists of a natural language prompt paired with a single image. The prompt is composed of two parts: (1) the original patient-authored query from the DermaVQA-IYII dataset, including both title and contextual description, and (2) a dermatology-specific assessment question derived from the Dermatology Assessment Schema (DAS). Unless otherwise specified, the task is framed as multiple-choice question answering with constrained outputs limited to the predefined DAS options. 

For cases involving multiple images per query, we apply task-specific aggregation strategies. For lesion location and size questions (e.g., CQID011-001 and CQID012-001), we assume each image depicts a single lesion and produce a combined prediction by taking the union of predicted (location, size) pairs across images. For lesion description tasks (e.g., CQID020-001), predictions from all images are aggregated as a union set. For color-related questions (e.g., CQID034-001), if at least two distinct colors are detected across images, we select option 8 ("combination (please specify)"). In all remaining cases, we select the option with the highest indexed label, excluding any “Not mentioned” options. This deterministic aggregation ensures consistent evaluation across models and image subsets.

We evaluate several state-of-the-art multimodal generative models, including Gemini-1.5-Pro, Gemini-2.0-flash, and Claude-3.7-Sonnet, GPT o4-mini, GPT-4.1, and o3. Each model is prompted using the combined question-image input, and outputs are post-processed using the same aggregation strategies described above.


\section{Segmentation Results}

Table \ref{tab:segmentation-res} summarizes segmentation performance under different prompting strategies and aggregation schemes. To leverage multiple gold standard masks for segmentation, a per-pixel majority vote was used to construct the gold standard for microscore calculations of the Jaccard and Dice indices used for ranking. In addition, the mean of the per-instance maximum scores and the mean of the per-instance mean scores across all test instances are also reported.

\begin{table*}[h]
  \centering
  \caption{Results of the closed-ended question answering task. 
  Best performance per category is shown in bold; second-best is underlined.} 
  \resizebox{\textwidth}{!}{
  \begin{tabular}{ c|ccccccccc|c }
        \hline
System & CQID010 & CQID011 & CQID012 & CQID015 & CQID020 & CQID025 & CQID034 & CQID035 & CQID036 & ALL \\ \hline

Gemini-1.5-Pro 
& 0.650 & \underline{0.876} & \underline{0.793} & 0.820 & 0.746 & 0.900 & \bf 0.660 & 0.890 & 0.710 & 0.783 \\

Gemini 2.0 Flash 
& 0.640 & 0.873 & 0.769 & 0.790 & \bf 0.757 & 0.930 & \underline{0.630} & 0.840 & 0.680 & 0.768 \\

Claude 3.7 Sonnet (2025-02-19) 
& 0.620 & 0.853 & 0.780 & \underline{0.840} & 0.720 & \underline{0.960} & 0.560 & 0.880 & 0.640 & 0.761 \\

o4-mini (2025-04-16) 
& \underline{0.660} & 0.847 & 0.590 & 0.810 & 0.735 & \underline{0.970} & 0.620 & \bf 0.900 & 0.430 & 0.729 \\

GPT-4.1 
& \bf 0.670 & 0.881 & \underline{0.830} & \bf 0.850 & \bf 0.757 & 0.930 & \underline{0.640} & 0.880 & \underline{0.730} & \underline{0.796} \\

o3 (2025-04-16) 
& 0.650 & \bf 0.893 & \bf 0.835 & 0.830 & \underline{0.756} & \bf 0.980 & 0.600 & \bf 0.900 & \bf 0.740 & \bf 0.798 \\

\hline
\end{tabular}
}
\label{tab:closedqa-res}
\end{table*}

Overall, BiomedParse consistently outperforms MedSAM, achieving higher Jaccard and Dice scores under all aggregation strategies.  Across both models, Mean of Max aggregation yields the strongest performance, followed by Mean of Mean, while Majority Vote results in comparatively lower scores. This pattern suggests that more permissive aggregation schemes (e.g., selecting the best overlap per instance) yield higher reported performance than strict pixel-level consensus evaluation. For BiomedParse, Prompt I achieves the best performance under Mean of Max and Mean of Mean aggregation, whereas Prompt III yields the highest Jaccard (0.3948) and Dice (0.5661) scores under the majority-vote-based microscore evaluation.

Overall, these findings underscore the importance of model choice and gold standard aggregation strategy, with performance differences across prompting strategies remaining model-dependent and sensitive to the evaluation aggregation scheme. 


\section{VQA Results}

Table \ref{tab:closedqa-res} reports the results of the closed-ended question answering task across nine question categories (CQID010-CQID036) and the micro-average across all questions (ALL).

 Because the same dermatological problem may have multiple sites, there may be related questions (e.g., "what is the size of the affected area for location 1", "what is the size of the affected area for location 2"). In these cases, the answers to the  related questions are collated together. Partial credit was given when there are partial matches to gold. For example, if gold annotations specify locations ``{head,neck}'', while system specifies ``{neck}'', a score of 0.5 is given. This evaluation design enables structured scoring while accounting for multi-site conditions\footnote{The evaluation code can be found here: \url{URL\_hidden\_for\_review}}. 

Overall, performance is strong across systems, with average accuracies ranging from 0.729 to 0.798. Among all evaluated models, o3 (2025-04-16) achieves the best overall performance (ALL = 0.798), closely followed by GPT-4.1 (0.796). The Gemini family exhibits competitive results, with Gemini-1.5-Pro outperforming Gemini 2.0 Flash on most categories and achieving an overall accuracy of 0.783.

Across individual question types, CQID025 (Itch) consistently yields the highest accuracies for all systems (>0.90), suggesting that binary symptom-based questions grounded in patient-reported information are comparatively easier. In contrast, CQID034 (Lesion Color) and CQID036 (Texture) show greater performance variability across models, indicating higher difficulty or ambiguity in fine-grained visual attribute recognition and controlled vocabulary selection. Notably, o4-mini shows strong performance on CQID025 (Itch) and CQID035 (Lesion Count) but underperforms on CQID012 (Size) and CQID036 (Texture), highlighting trade-offs between model compactness and robustness on spatially grounded and visually nuanced question types.

To study the affect of multiple site problems (a proxy for quesiton complexity), we calculate the correlation of between number of site locations with question accuracy. To normalize for differences in site-size paired slots and site-description paired slots, we modified accuracy to check against the paired accuracy from the gold. The result spearman correlations are shown in Table \ref{tab:probcompl}. While correlations with most question types are small or negligible, we observe some comparatively higher negative correlations with size and description accuracy. This is consistent with the idea that the more sites exists, the more difficult it is to get the corresponding sizes and descriptions all correct.

\begin{table}[h]
\centering
\caption{Correlation Between Site Number and Accuracy}
\label{tab:probcompl}
\small
\begin{tabular}{ccc}
\hline
QID & Spearman & Desc \\
\hline
CQID010 & -0.147 & Site Extent\\
CQID015 & 0.116 & Onset\\
CQID025 & 0.116 & Itch\\
CQID034 & 0.134 & Lesion Color\\
CQID035 & -0.001 & Lesion Count\\
CQID036 & 0.046 & Texture\\
\hline
CQID011-CQID012 & -0.461 & Location-Size\\
CQID011-CQID020 & -0.394 & Location-Desc\\
\hline
\end{tabular}
\end{table}

\section{Conclusion}

While segmentation techniques can provide critical information for reasoning about critical dermatology questions, e.g. number of lesions. The masks can moreover be further processed down-stream for less direct information such as skin lesion color and texture. However, we see that the base performances of MedSAM and BiomedParse provide promising but modest performances. These may be attributable to the highly challenging nature of consumer supplied dermatology images where subject focus, angle, lighting, etc can vary widely.  In the closed-ended question answering task, we observe strong performance for some individual questions, such as itch, whereas questions requiring fine-grained visual interpretation, such as lesion texture, remain more challenging. These results demonstrate that structured dermatological assessment encompasses attributes with substantially different levels of difficulty and highlight persistent challenges in extracting fine-grained clinical information from patient-generated data. Future work can improve segmentation and leverage its outputs for specific question answering tasks, for example through models that integrate spatial and visual features to characterize lesion number, color, texture, and shape.

We present DermaVQA-DAS, a substantial extension of the DermaVQA dataset that introduces closed-ended question answering and dermatological lesion segmentation as new benchmark tasks. Central to this contribution is the Dermatology Assessment Schema (DAS), a novel, expert-developed framework that systematically encodes clinically relevant dermatological features in a structured and interpretable form. By releasing bilingual assessment questions and expert annotations, DermaVQA-DAS supports research in structured clinical reasoning, multimodal understanding, and cross-lingual dermatological AI.

Through comprehensive benchmarking of several multimodal models, we establish baselines for both closed-ended QA and segmentation. More broadly, DermaVQA-DAS provides a framework for evaluating the transformation of unstructured dermatological encounters into structured, clinically meaningful representations, while enabling systematic assessment of performance across distinct dermatological attributes. Importantly, as the schema was designed to be both clinically relevant and patient facing, this schema can be adopted into AI facing technology that directly gets user-input from patients before forwarding to relevant clinicians. DermaVQA-DAS has been adopted in an international challenge (Reference\_hidden\_for\_review), underscoring its relevance as a community benchmark. We publicly release the dataset and the DAS schema to facilitate reproducibility and support the development of clinically grounded dermatological AI systems.  

\section*{Limitations} 
Despite its contributions, this work has several limitations. First, although DermaVQA-DAS expands beyond dermatoscopic-only benchmarks, the dataset size and clinical diversity remain constrained relative to real-world dermatology practice, which may limit generalization across rare conditions, skin tones, and imaging settings. Second, the Dermatology Assessment Schema, while expert-designed and bilingual, reflects a fixed set of predefined attributes and may not fully capture open-ended patient concerns or nuanced clinical reasoning used by dermatologists.

\section*{Ethical Considerations}

Internally hired medical annotators were compensated at fair wage rates based on their state of residence, while physician expertise was provided through academic collaborations without monetary compensation. We recognize that the potential benefits of this work may not extend equally to all populations due to biases or imbalances in the underlying data and disparities in access to technology. Licenses for all underlying datasets were reviewed for compatibility with the intended use case. The release of the new dataset extensions was assessed for compliance with our organization’s policies, and applicable data use terms are published in the data repository. The technology developed in this work is intended to be used with appropriate medical supervision, and unequal access to such supervision may further limit its benefits for some populations.

\section*{Acknowledgements}

This study received no funding.


\bibliography{main}

\appendix


\end{document}